\documentclass[conference]{IEEEtran}
\IEEEoverridecommandlockouts

\usepackage{cite}
\usepackage{amsmath,amssymb,amsfonts}
\usepackage{algorithmic}
\usepackage{color}
\usepackage{graphicx}
\usepackage{subcaption}

\usepackage{tcolorbox}
\usepackage{textcomp}
\usepackage{xcolor}
\usepackage{booktabs}
\usepackage{tikz}
\usepackage[bottom]{footmisc}
\def\BibTeX{{\rm B\kern-.05em{\sc i\kern-.025em b}\kern-.08em
    T\kern-.1667em\lower.7ex\hbox{E}\kern-.125emX}}
    
\makeatletter
\def\ps@IEEEtitlepagestyle{%
  \def\@oddfoot{\mycopyrightnotice}%
}
\def\mycopyrightnotice{%
\fbox{\parbox{\dimexpr\textwidth\fboxsep\fboxrule\relax}{
\begin{minipage}{\textwidth\fboxsep\fboxrule}
  \footnotesize
  \textcopyright 2022 IEEE. Personal use of this material is permitted. Permission from IEEE must be obtained for all other uses, in any current or future media, including reprinting/republishing this material for advertising or promotional purposes, creating new collective works, for resale or redistribution to servers or lists, or reuse of any copyrighted component of this work in other works.
  \end{minipage}
}}
}
\makeatother

\begin{document}

\title{A Linguistic Investigation of Machine Learning based Contradiction Detection Models: An Empirical Analysis and Future Perspectives\thanks{This research will be published in the 2022 proceedings of the IEEE International Conference on Machine Learning and Applications (ICMLA) and has been funded by the Federal Ministry of Education and Research of Germany and the state of North-Rhine Westphalia as part of the Lamarr-Institute for Machine Learning and Artificial Intelligence, LAMARR22B.}
}

\author{
    \IEEEauthorblockN{ Maren Pielka\IEEEauthorrefmark{2}, Felix Rode\IEEEauthorrefmark{1}\IEEEauthorrefmark{2}, Lisa Pucknat\IEEEauthorrefmark{1}\IEEEauthorrefmark{2},
        Tobias Deußer\IEEEauthorrefmark{1}\IEEEauthorrefmark{2}, Rafet Sifa\IEEEauthorrefmark{2}
    }
    \IEEEauthorblockA{\IEEEauthorrefmark{1} University of Bonn, Bonn, Germany}
    \IEEEauthorblockA{\IEEEauthorrefmark{2} Fraunhofer IAIS, Sankt Augustin, Germany}
    \texttt{Maren.Pielka@iais.fraunhofer.de}
    \\ 
}

\maketitle

\begin{abstract}
  We analyze two Natural Language Inference data sets with respect to their linguistic features. The goal is to identify those syntactic and semantic properties that are particularly hard to comprehend for a machine learning model. To this end, we also investigate the differences between a crowd-sourced, machine-translated data set (SNLI) and a collection of text pairs from internet sources. Our main findings are, that the model has difficulty recognizing the semantic importance of prepositions and verbs, emphasizing the importance of linguistically aware pre-training tasks. Furthermore, it often does not comprehend antonyms and homonyms, especially if those are depending on the context. Incomplete sentences are another problem, as well as longer paragraphs and rare words or phrases. The study shows that automated language understanding requires a more informed approach, utilizing as much external knowledge as possible throughout the training process.
\end{abstract}

\maketitle

\begin{table*}[h]
\centering
\small
\begin{tabular}{ccc} \toprule
    \textbf{Premise} & \textbf{Hypothesis} & \textbf{Label} \\ \midrule
    \begin{minipage}{5cm} \vspace{1mm} "Eine Person auf einem Pferd springt über ein zusammengebrochenes Flugzeug." - \textit{``A person on a horse jumps over a broken down airplane."} \vspace{1mm} \end{minipage} & \begin{minipage}{5cm} \vspace{1mm} "Eine Person trainiert ihr Pferd für ein Turnier."- \textit{``A person is training his horse for a competition."} \vspace{2mm} \end{minipage} & "no contradiction" \\ \midrule
     \begin{minipage}{5cm} \vspace{1mm} "Kinder lächeln und winken der Kamera zu." - \textit{``Children smiling and waving at camera"} \vspace{1mm} \end{minipage} & \begin{minipage}{5cm} \vspace{1mm} "Die Kinder runzeln die Stirn." - \textit{``The kids are frowning"} \vspace{1mm} \end{minipage} & "contradiction" \\ 
      \bottomrule
      \end{tabular}
\caption{Examples from the SNLI data set, machine-translated German version and English original (in italic)}
\label{tab:snli_examples}
\end{table*}

\section{Introduction}
Contradiction Detection and Natural Language Inference are particularly hard problems from the Natural Language processing (NLP) domain \cite{cidm2019, icpr2021, pucknat2021detecting}. A variety of Machine Learning (ML) approaches have been introduced to tackle this task, the state-of-the-art being transformer-based methods such as XLM-RoBERTa \cite{liu_roberta, conneau2019unsupervised}. While achieving overall good classification performance, those methods still lack understanding of linguistic features, and are relying heavily on extensive amounts of data for language model pre-training. To this end, we investigate the well-known SNLI data set \cite{Bowman2015-EMNLP} with the aim to discover distinct linguistic properties that are important in recognizing contradictions. In addition, we also collect a data set in German language from various online sources, which is being labeled by human annotators for the contradiction detection task. The objectives of this work are two-fold:
\begin{itemize}
    \item We want to find out, whether the types of contradictions differ between the synthetically created SNLI data, and the data we collected from online sources. This will also help us building a model that works well in a real-world application scenario.
    \item We want to identify relevant linguistic features, that could help an ML model learn to recognize contradictions, without relying on extensive amounts of pre-training data. To this end, we analyze the predictions of an XLM-RoBERTa model with respect to those syntactic and semantic features that lead to wrong classifications.
    
\end{itemize}
Based on those findings, we develop some first ideas for informed ML approaches that would help improve those results in the future, by injecting linguistic knowledge into the learning process. Our analysis is limited to the German language, but many of the results could most likely be applied to other languages as well.

\section{Related Work}
There is little prior work on the subject of linguistically aware modelling for Natural Language Inference. Marneffe et al. \cite{Marneffe2008-ACL} are the first to comprehensively define Contradiction Detection as a distinct problem, and present some initial methods based on handcrafted semantic features such as antonymity and polarity. They point out, that detecting contradictions is a particularly hard task, because a deep level of language understanding is needed, as well as some background knowledge that cannot necessarily be inferred from the analyzed text alone. 

Li et al. \cite{Li2017} attempt to learn contradiction-specific word embeddings by enforcing words with opposite meaning to be mapped into different regions of feature space. This addresses the issue, that opposite words (antonyms) tend to appear in similar contexts, such that a conventional word embedding model would learn similar embeddings for those. Using this method, the authors report state-of-the-art results on the SemEval task.

A more recent approach is presented by \cite{limit_bert2019}. They introduce a novel, linguistically aware combination of pre-training objectives for BERT \cite{devlin2018bert}, including POS-tagging, semantic role labeling and syntactic parsing. The idea is for the model to capture a high level of semantic knowledge which will be helpful in fine-tuning on a downstream task. They achieve competitive results on the GLUE benchmark.

Pucknat et al. \cite{pucknat2021detecting} evaluated different neural network based approaches on the Contradiction Detection task in German language. While XLM-RoBERTa \cite{liu_roberta, conneau2019unsupervised} achieves the best results of all models under investigation, it still has problems with complicated syntactic structures and real-world language use. This gave rise to the idea of investigating the role of linguistic features more closely, in order to come up with an informed learning approach.

\section{Data}
For our analysis, we use two different data sources: A translated version of the Stanford Natural Language Inference (SNLI) data set, and a collection of real-world examples from various online sources in German language.

\subsection{SNLI}
The Stanford Natural Language Inference (SNLI) data set was first introduced by Bowman et al. \cite{Bowman2015-EMNLP}. It is the largest collection of human-generated premise and hypothesis pairs for the NLI task to date, with over 570,000 examples. The data was collected in a crowd-source campaign, where both samples and labels were created by human annotators. The final labels were decided upon by a majority vote, thus minimizing noise due to human error and ambiguity. In the original data set, there are three possible labels: \textit{entailment}, \textit{neutral} and \textit{contradiction}. For the purpose of the Contradiction Detection objective, we binarize those labels by consolidating the \textit{neutral} and \textit{entailment} labels to \textit{no contradiction}.

A large portion of the SNLI training set (100 000 examples), as well as the whole validation and test set were machine-translated to German \cite{cidm2019}, using the DeepL API\footnote{ https://github.com/fraunhofer-iais/snli\_translated}. The data set was found to be of overall sufficient quality, but there are some artifacts and inconsistencies, due to the machine translation and the annotation setup. Because of those issues, it is not completely representative of a real-world setting.

\subsection{Online Data Set}
To address those shortcomings, we collected a data set from various online sources in German language\footnote{The data set will be made publicly available upon the acceptance of this work.}. Those include news \footnote{https://correctiv.org, https://nachrichtenleicht.de/}, tweets \footnote{https://twitter.com/}, company and employer ratings \footnote{https://de.trustpilot.com/, https://www.kununu.com/}, game reviews  \footnote{https://store.steampowered.com/} and product reviews \footnote{https://www.amazon.de/}.

The data is being manually annotated by six workers using two different modes. For the first one, the annotators are presented with random examples from all five sources, and shall come up with contradicting or non-contradicting hypotheses for each of those examples. There is also the option to exclude sentences, if no meaningful hypothesis can be found. Since this procedure is quite costly, we additionally use another annotation mode, presenting the annotators with pairs of sentences from the online sources, where the premise and hypothesis have already been matched. To achieve this, the samples are first being grouped into different categories, according to the meta-data from the website (e.g. similar keywords on Twitter). Additionally, a text similarity measure is applied to identify samples that are likely to refer to the same topic. Those text pieces that belong to the same category, and show a high similarity are then being matched and presented to the reviewers as premise-hypothesis pairs. Given this setup, the annotators only have to add the respective label: "contradiction", "no contradiction" or "exclude" (for cases where the sentences do not relate to each other, or one of them makes no sense).

We create 10 000 data points using the first annotation procedure, and another 31 000 using the second approach. The 10 000 manually created examples are being reviewed by a second annotator, to minimize noise due to subjectivity and human error. After those steps, 531 samples had to be excluded, so that we end up with a total of 40 589 examples. A random 60-20-20 training-validation-test split is being applied to the remaining data set. 

\section{Linguistic Analysis}
We perform a linguistic analysis of the two data sets, focussing on the qualitative differences between SNLI and internet data, and those instances that impose problems for the classifier. For this evaluation, an XLM-RoBERTa model \cite{liu_roberta, conneau2019unsupervised} is used, which has been pre-trained for the Masked Language Modeling task on 100 languages, and fine-tuned on the respective training set for the Contradiction Detection task (translated SNLI / online data). For details on the architecture and training procedure, we refer to \cite{pucknat2021detecting}, as the focus of this paper is an analytical one.

\subsection{Differences between the two Data Sets}
The two analyzed data sets differ primarily in syntactic structure. The first (SNLI) has basic syntactic structures and semantics as well as grammatical simplicity, whereas the second data set (internet data) virtually lives from syntactic versatility. Here, the data sets differ not only in sentence-related verbosity and sentence length but also in the juxtaposition of such sentences. The sentences from the internet data are also formulated more homogeneously and thus come much closer to real language use. Some example sentence pairs from the two data sets can be found in tables \ref{tab:snli_examples} and \ref{tab:internet_examples}.

\begin{table*}[h]
\centering
\small

\begin{tabular}{ccc} \toprule
    \textbf{Premise} & \textbf{Hypothesis} & \textbf{Label} \\ \midrule
    \begin{minipage}{5cm} \vspace{1mm} "Die Qualität der Kette ist sehr gut. Die Kette sieht hochwertig aus und die Lieferung war wirklich schnell :)" - \textit{``The quality of the necklace is very good. The necklace looks high-quality and delivery was really fast :)"} \vspace{1mm} \end{minipage} & \begin{minipage}{5cm} \vspace{1mm} "Keine Benachrichtigung über Sendung und keine sendungsverfolgung möglich.  Zu lange Lieferzeiten."- \textit{``No notification on the shipment and shipment tracking not possible. Delivery times too long."} \vspace{2mm} \end{minipage} & "contradiction" \\ \midrule
     \begin{minipage}{5cm} \vspace{1mm} "Das Unternehmen schreibt sich das Thema hoch auf die Fahne. Leider steht es nur da.Angefangen von der Mülltrennung bis zum Versand von E-Teilen, die in überdimensionierten Kartons versendet werden." - \textit{``The company claims to highly prioritize the topic. Unfortunately, that is all it does. Starting with waste separation, as well as mailing spare parts in oversized boxes."} \vspace{1mm} \end{minipage} & \begin{minipage}{5cm} \vspace{1mm} "Es werden gerne mal zum CSD etc. Marketingaktionen gestartet oder das T-Logo in Regenbogenfarben angemalt. Nachhaltig ist das aus meiner Sicht nicht" - \textit{``It is common for the company to start marketing campaigns on the occasion of CSD etc., or paint the T-logo in rainbow colors. From my point of view, none of this is sustainable."} \vspace{1mm} \end{minipage} & "no contradiction" \\ 
      \bottomrule
      \end{tabular}
\caption{Examples from the internet data set, original German version and English translation (in italic), with binarized labels}
\label{tab:internet_examples}
\end{table*}

\subsection{Challenges for the NLI model}
The model put out faulty analyses whenever it was confronted with grammatically incomplete and incorrect sentences. As soon as one of the sentences showed grammatical deficiencies in the form of sentence breaks (anacoluth) or word cuts, problems arose with the recognition  of the reference word or the sentence’s meaning. Another area of concern is the record length. The model often failed to recognize the syntactic and semantic keywords/signifiers when confronted with longer and more complicated sentences.

\begin{tcolorbox}[notitle,boxrule=0pt,boxsep=0pt,left=1em,right=1em,
top=0.5em,bottom=0.5em,colback=gray!10,colframe=gray!10, fontupper=\color{darkgray}]
\textbf{Premise}: ``Ich finde den Ansatz mit den Bioölen sehr gut. Deshalb habe ich mich auch für eine Bestellung entschieden." $-$ \textit{``I think the approach with the organic oils is very good. That is why i decided to place an order"}
\textbf{Hypothesis:} ``Ich habe selten so viel Kulanz und Entgegenkommen von einem Händler erlebt. Ich habe am dritten Februar bestellt, leider kam das Paket nicht zum angezeigten Liefertermin. Ein kurzer E-Mail Kontakt mit dem Händler zeigte, dass das Paket beim Zusteller verloren gegangen war. Ohne Umstände wurde sofort ein neues Paket  losgeschickt, was auch 3 Tage später ankam doch leider war dort eine Flasche kaputt. Ein erneuter E-Mail Kontakt und schon wurde die Flasche ersetzt, aber nicht nur das mir wurde auch noch eine Flasche, wegen den ganzen Umständen geschenkt." \textit{``I have rarely experienced so much goodwill and responsiveness from a retailer. I ordered on 3 February, but unfortunately the package did not arrive on the date indicated. A brief email contact with the retailer showed that the parcel had been lost by the delivery company. A new parcel was immediately sent, which arrived three days later, but unfortunately one of the bottles was broken. Another email contact and the bottle was replaced, but not only that, I was also given a bottle as a gift because of all the circumstances."}\vspace{2mm} \\ 
\textbf{Gold label:} No contradiction \\
\textbf{Prediction}: Contradiction
\end{tcolorbox}

In this example, a grammatical construction consisting of two main clauses is juxtaposed with a multi-membered construction, in this case consisting of main and subordinate clauses. The syntactically and semantically important signifier of the premise "with the bio-oils" is suppressed in the hypothesis and not explicitly emphasised again. Even if there is a clear connection in terms of content, it is lost in the stringing together of individual semantic hierarchies. A clear semantic analysis is hardly possible for the model in this form due to this structure and the lack of reference words or similarities.

The following example also shows problems with the accumulation of sentences in juxtaposition to short or even elliptical sentences. The meaning and structure of the conditional construction is radically changed in the hypothesis. Furthermore, it is questionable whether the meaning of such short phrases as "fits exactly" can really be determined and related. The same problem naturally occurs with grammatically incomplete sentences. Missing reference words and inter-syntactic relations cannot be sufficiently captured, even though these sentence constructions are accepted in both German and English and may fall under the genre of linguistic devices.

\begin{tcolorbox}[notitle,boxrule=0pt,boxsep=0pt,left=1em,right=1em,
top=0.5em,bottom=0.5em,colback=gray!10,colframe=gray!10, fontupper=\color{darkgray}]
\textbf{Premise}: ``Ja okay, Basic heißt nicht hochwertig. Auf
dieses Laken möchte man sich nun wirklich nicht legen.
Das Laken greift sich sehr unangenehm ist eigentlich kaum zu beschreiben, ähnlich Plastik. Ich hatte dieses vorgesehen für meinen Mieter einer möblierten Wohnung, aber ich denke das möchte ich ihm nicht zumuten." $-$ \textit{``Yes okay, basic does not mean high quality. You really don't want to lie on this sheet. The sheet is very unpleasant to the touch, it's hard to describe, similar to plastic I had intended this for my tenant in a furnished flat, but I don't think I want to put him through that."}
\textbf{Hypothesis:} ``Passt genau!" \textit{``Fits perfectly!"}\vspace{2mm} \\ 
\textbf{Gold label:} No contradiction \\
\textbf{Prediction}: Contradiction
\end{tcolorbox}

In addition the model struggled with negations. If a negation was not directly related to the signifier, flawed results were produced. To this effect, there have been cases in which additional words could stand out and shift the meaning. Furthermore, the model did not deal well with technical or rare terms and was unable to compare them adequately. At the same time ambiguity, antonyms, homonyms and homonymous verbs were not recognized and could therefore not be linked correctly. The model seemed to recognize metaphors and allegories only to a limited extent. The previously mentioned uncertainty with individual terms, as well as the semantic capturing of individual parts of sentences, lead
to most errors in terms of recognizing contradictions. 

\begin{tcolorbox}[notitle,boxrule=0pt,boxsep=0pt,left=1em,right=1em,
top=0.5em,bottom=0.5em,colback=gray!10,colframe=gray!10, fontupper=\color{darkgray}]
\textbf{Premise}: ``Auf der Demonstration hatten die Demonstranten mit viel Rauch und Nebel zu kämpfen." $-$ \textit{``On the demonstration, the demonstrators had to deal with a lot of smoke and fog"}\vspace{2mm} \\ 
\textbf{Hypothesis:} ``Die Polizei setzte Tränengas gegen die Demonstranten ein." \textit{``The police used tear gas against the demonstrators."}\vspace{2mm} \\ 
\textbf{Gold label:} No contradiction \\
\textbf{Prediction}: Contradiction
\end{tcolorbox}

This example shows another interesting problem of the model. The premise receives the state description of an environmental occurrence. "smoke and fog" stand together here as a methaporic synonym to the statement of "tear gas" contained in the hypothesis. The model cannot recognize the metaphor. The same applies to homonyms and homonymous verbs. Once the meanings are not presented in a direct way, the model cannot analyze possible contradictions due to lack of understanding. Furthermore, such word types can often only be evaluated from the context. Thus, this bivalent problem appears to be a great challenge for the model.

The greatest error rate, however, was seen in the analysis and assignment of local prepositions. These could only rarely or not at all be distinguished from one another and were treated in the same way by the model although they fulfil a major semantic function.

\begin{tcolorbox}[notitle,boxrule=0pt,boxsep=0pt,left=1em,right=1em,
top=0.5em,bottom=0.5em,colback=gray!10,colframe=gray!10, fontupper=\color{darkgray}]
\textbf{Premise}: ``Tim sitzt neben der Badewanne." $-$ \textit{``Tim sits next to the bathtub."}\vspace{2mm} \\
\textbf{Hypothesis:} ``Tim wäscht sich." \textit{``Tim washes himself."}\vspace{2mm} \\ 
\textbf{Gold label:} Contradiction \\
\textbf{Prediction}: No contradiction
\end{tcolorbox}

This example was not part of one of the data sets, but was created by us to test the ability of the model to recognize semantic differences when the sentence is being slightly altered. We experimented with replacing the respective local preposition by other local prepositions ("in/auf/neben/unter"), which, however, give a completely different semantic implication. We wanted to test whether the model's prediction would change, but it yielded the same result as the original sentence. Furthermore, we added a subordinate clause (``... während eine Frau sich wäscht" / ``while a woman is washing herself") that directs the meaning to another object and yet the model stuck to an incorrect analysis and made no distinctions among the prepositions or the sentence content. It was interesting that these errors did not occur with other preposition types.

\section{Conclusion and Future Work}
We presented a first comprehensive, in-depth linguistic analysis of two data sets for the Contradiction Detection task in German language. In doing so, we discovered a number of syntactic and semantic features that pose a challenge to the transformer model. These valuable insights open a variety of opportunities for future research. One possibility would be to define additional pre-training tasks that capture syntactic and semantic knowledge. For example, part of speech tagging could be helpful, given that the model has problems with prepositional references and missing verbs. It could also be worthwhile exploiting semantic webs like WordNet or GermaNet to gain knowledge about antonyms and synonyms.

Another option is to use formalized knowledge to pre-train the model, as described by \cite{laura_informed}. This could be achieved, by constructing characteristic, simplified training examples for types of contradictions that the model struggles with, based on the linguistic knowledge we gathered. Those could also be produced in a semi-automated fashion, by applying rules inferred from the external knowledge. 
Given that the current model has difficulties capturing the semantics of metaphors and rare terms, it would most likely be worthwhile to look into domain-specific fine-tuning methods. For example, in the context of financial textual data analysis \cite{sifa2019towards, hillebrand2022kpi} information-extraction could be applied as a pre-training task.

Finally, it would also be interesting to validate our findings by conducting a quantitative analysis, namely to test in how many cases the classifier gives an incorrect prediction for a sample that shows a specific linguistic phenomenon.

\section*{Acknowledgements}
We would like to thank Pascal Binias, Moritz Haidl, Torben Hoffmann, Matilda Heinen, Lars Otte and Henning Otte for manually annotating the internet data set.

\bibliography{custom}
\bibliographystyle{plain}

\end{document}